\documentclass[11pt,a4paper]{article}
\usepackage[hyperref]{aacl-ijcnlp2020}

\usepackage{times}
\usepackage{latexsym}

\usepackage{enumitem}
\usepackage{graphicx}
\usepackage{amsmath}
\usepackage{amssymb}
\usepackage{multirow}
\usepackage{tablefootnote}

\usepackage{microtype}

\aclfinalcopy 
\def\aclpaperid{264} 

\title{Modality-Transferable Emotion Embeddings for Low-Resource Multimodal Emotion Recognition}

\author{Wenliang Dai, Zihan Liu, Tiezheng Yu, Pascale Fung \\
Center for Artificial Intelligence Research (CAiRE)\\
Department of Electronic and Computer Engineering\\
The Hong Kong University of Science and Technology, Clear Water Bay, Hong Kong\\
\texttt{\{wdaiai,zliucr,tyuah\}@connect.ust.hk, pascale@ece.ust.hk}}

\date{}

\begin{document}
\maketitle
\begin{abstract}
Despite the recent achievements made in the multi-modal emotion recognition task, two problems still exist and have not been well investigated: 1) the relationship between different emotion categories are not utilized, which leads to sub-optimal performance; and 2) current models fail to cope well with low-resource emotions, especially for unseen emotions. In this paper, we propose a modality-transferable model with emotion embeddings to tackle the aforementioned issues. We use pre-trained word embeddings to represent emotion categories for textual data. Then, two mapping functions are learned to transfer these embeddings into visual and acoustic spaces. For each modality, the model calculates the representation distance between the input sequence and target emotions and makes predictions based on the distances. By doing so, our model can directly adapt to the unseen emotions in any modality since we have their pre-trained embeddings and modality mapping functions. Experiments show that our model achieves state-of-the-art performance on most of the emotion categories. Besides, our model also outperforms existing baselines in the zero-shot and few-shot scenarios for unseen emotions \footnote{Code is available at \url{https://github.com/wenliangdai/Modality-Transferable-MER}}.
\end{abstract}

\section{Introduction}
\label{sec:introduction}
\begin{figure}[t!]
    \centering
    \includegraphics[width=\linewidth]{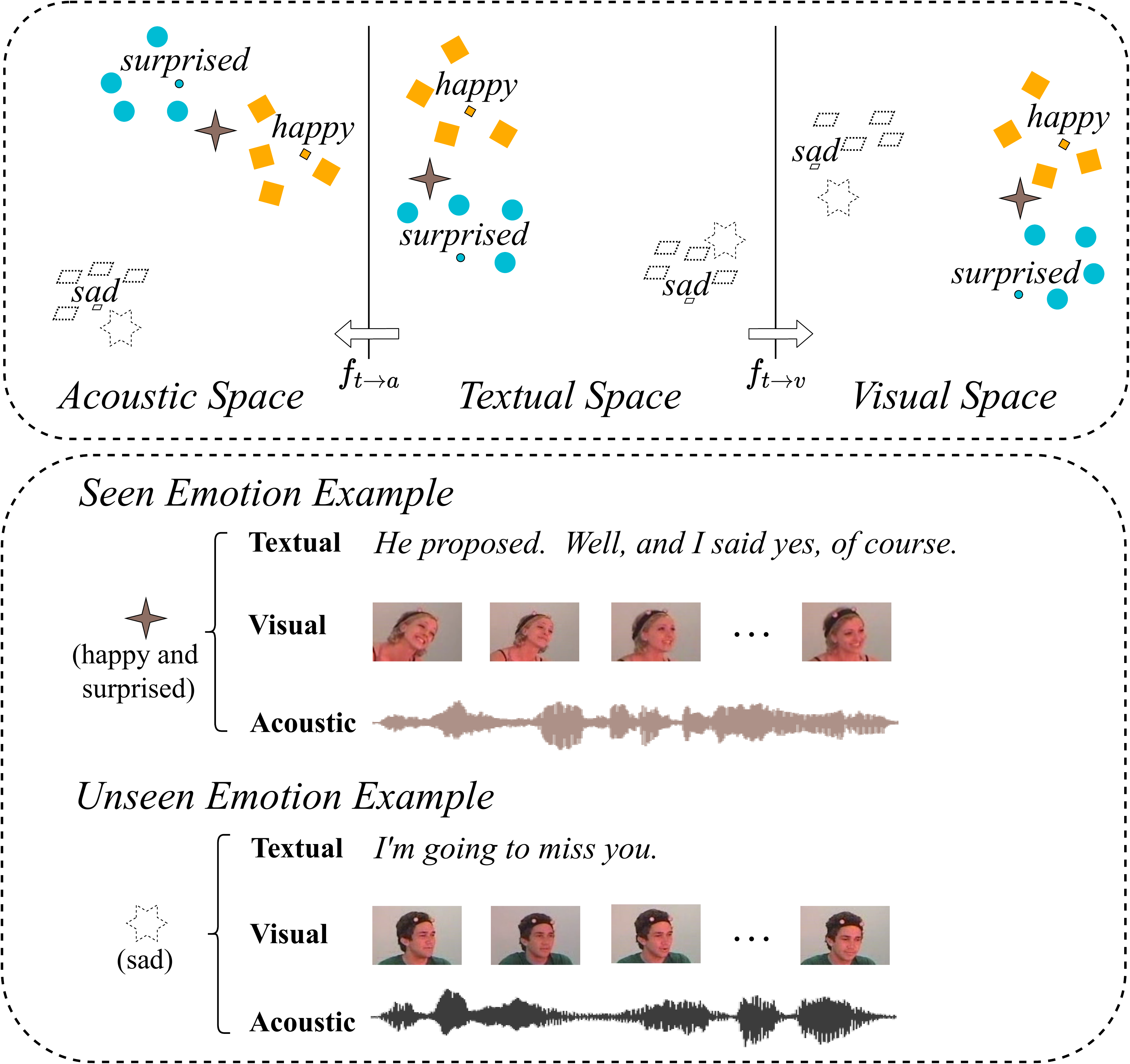}
    \caption{An intuitive example of our method. In the upper image, the relative positions of GloVe emotion embeddings (\textit{happy}, \textit{surprised}) are shown in the textual space, which are then projected to acoustic and visual spaces by two mapping functions ($f_{t \rightarrow a}$ and $f_{t \rightarrow v}$). Our model learns to group the representations of input sentences (\includegraphics[height=\fontcharht\font`\B]{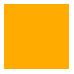}, \includegraphics[height=\fontcharht\font`\B]{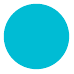}) based on their corresponding emotion embeddings. Examples are shown in the lower image. When a sample has both \textit{happy} and \textit{surprised} emotions, its representation gets close to these two emotion embeddings in all three spaces. If an unseen emotion \textit{sad} (\includegraphics[height=\fontcharht\font`\n]{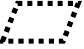}) comes, the model processes it with the same flow and recognizes corresponding data samples.}
    \label{fig:intro}
\end{figure}

Multi-modal emotion recognition is an increasingly popular but challenging task. One main challenge is that labelled data is difficult to come by as humans find it time-consuming to discern emotion categories from either speech or video. Indeed we humans express emotions through a combination of modalities, including the way we speak, the words we use, facial expressions and sometimes gestures. It is also much more comfortable for humans to understand each other's emotions when they can both hear and see the other person. It follows that multi-modal emotion recognition can, therefore, yield more reliable results than restricting machines to a single modality. 

In the past few years, much research has been done to better understand intra-modality and inter-modality dynamics, and modality fusion is a widely studied approach. For example, \citet{tensor-fusion-network-2017} proposed a tensor fusion network that combines three modalities from vectors to a tensor using the Cartesian product. In addition, the attention mechanism is commonly used to do modality fusion~\citep{memory-fusion-network, RAVEN, rmfn, CMN, found-in-translation, MulT2019}. Although significant improvements have been made on the multi-modal emotion recognition task, however, the relationship between emotions has not been well modelled, which can lead to sub-optimal performance. Also, the problem of low-resource multi-modal emotion recognition is not adequately studied. Multi-modal emotion recognition data is hard to collect and annotate, especially for low-resource emotions (e.g., surprise) that are rarely seen in daily life, which motivates us to investigate this problem. 

In this paper, we propose a modality-transferable network with cross-modality emotion embeddings to model the relationship between emotions. Given that emotion embeddings contain semantic information and emotion relations in the vector space, we use them to represent emotion categories and measure the similarity of the representations between the input sentence and target emotions to make predictions. Concretely, for the textual modality, we use the pre-trained GloVe~\citep{glove} embeddings of emotion words as the emotion embeddings. As there are no pre-trained emotion embeddings for the visual and acoustic modalities, the model learns two mapping functions, \(f_{t \rightarrow v}\) and \(f_{t \rightarrow a}\), to transfer the emotion embeddings from the textual space to the visual and acoustic spaces (Figure \ref{fig:intro}). Therefore, for each modality, there will be a dedicated set of emotion embeddings. The distances computed in all modalities will be finally fused, and the model will make predictions based on that.

Benefiting from this prediction mechanism, our model can easily carry out zero-shot learning (ZSL) to identify unseen emotion categories using the embeddings from unseen emotions. The intuition behind it is that the pre-trained and projected emotion embeddings form a semantic knowledge space, which is shared by both the seen and unseen classes. Furthermore, with the help of embedding mapping functions, the model can also perform ZSL on a single modality during inference time. When a few samples from unseen emotions are available, our model can adapt to new emotions without forgetting the previous emotions by using joint training and continual learning~\citep{lopez2017gradient}.

Our contributions in this work are three-fold:

\begin{itemize}
    \item We introduce a simple but effective end-to-end model for the multi-modal emotion recognition task. It learns the relationship of different emotion categories using emotion embeddings.
    \item To the best of our knowledge, this paper is the first to investigate multi-modal emotion recognition in the low-resource scenario. Our model can directly adapt to an unseen emotion, even if only one modality is available.
    \item Experimental results show that our model achieves state-of-the-art results on most emotion categories. We also provide a thorough analysis of zero-shot and few-shot learning.
\end{itemize}

\begin{figure*}[t!]
    \centering
    \includegraphics[width=\textwidth]{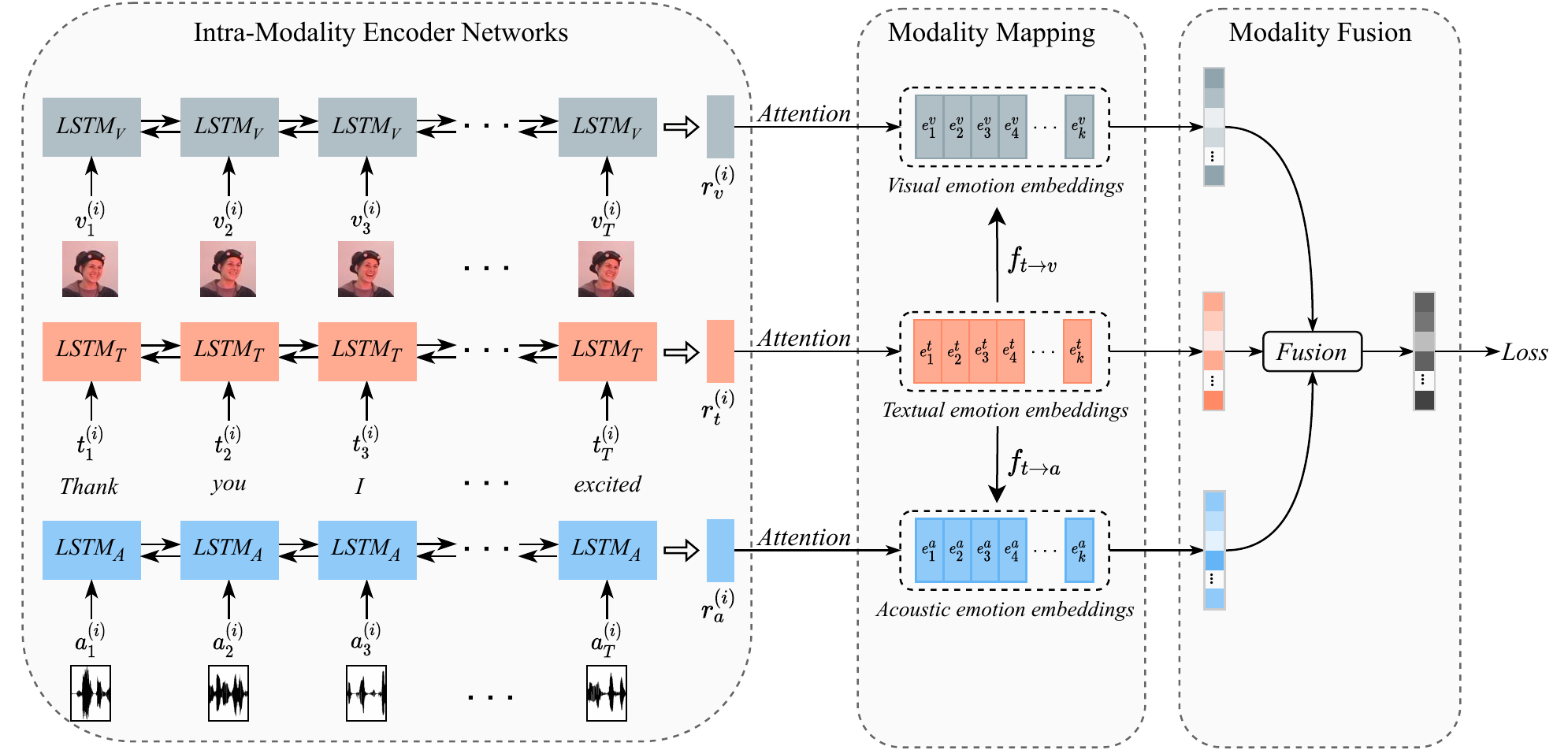}
    \caption{The architecture of our proposed multi-modal emotion recognition model. It consists of three LSTM networks, one emotion embedding mapping module, and one modality fusion module. For each modality, the input is a sequence of length $T$. Each modality has a set of emotion embeddings by mapping the GloVe textual emotion embeddings to the other modalities using \(f_{t \rightarrow v}\) and \(f_{t \rightarrow a}\). The whole architecture is optimized end-to-end.}
    \label{fig:model}
\end{figure*}

\section{Related Works} \label{sec:related_works}

\subsection{Multi-modal Emotion Recognition}
Since the early 2010s, multi-modal emotion recognition has drawn more and more attention with the rise of deep learning and its advances in computer vision and natural language processing~\citep{baltruvsaitis2018multimodal}. \citet{avec2011} proposed the first Audio-Visual Emotion Challenge and Workshop (AVEC), which focused on multi-modal emotion analysis for health. In recent years, most achievements in this area aimed to find a better modality fusion method. \citet{tensor-fusion-network-2017} introduced a tensor fusion network that combined data representation from each modality to a tensor by performing the Cartesian product. In addition, the attention mechanism~\citep{Bahdanau2015NeuralMT} has been widely applied to do modality fusion and emphasis~\citep{memory-fusion-network,found-in-translation,MulT2019}. Furthermore, \citet{efficient-low-rank-2018} proposed a low-rank architecture to decrease the problem complexity, and \citet{factorized-multimodal} introduced a modality re-construction method to generate occasional missing data in a modality. 

Although prior works have made progress on this task, the relationship between emotion categories has not been well modelled in previous works, except by \citet{xu2020emograph}, who captured emotion correlations using graph networks for emotion recognition. However, the model is only based on a single textual modality. Additionally, the previous studies have not put much effort toward unseen and low-resource emotion categories, which is a problem of multi-modal emotion data by nature.

\subsection{Zero/Few-Shot and Continual Learning}
Zero-shot and few-shot learning methods, which address the data scarcity scenario, have been applied to many popular machine learning tasks where zero or only a few training samples are available for the target tasks or domains, such as machine translation~\citep{johnson2017google,gu2018meta}, dialogue generation~\citep{zhao2018zero,madotto2019personalizing}, dialogue state tracking~\citep{liu2019attention,wu2019transferable}, slot filling~\citep{bapna2017towards,liu2019zero,liu2020coach}, and accented speech recognition~\citep{winata2020learning}. They have also been adopted in multiple cross-lingual tasks, such as named entity recognition~\citep{xie2018neural,ni2017weakly}, part-of-speech tagging~\citep{wisniewski2014cross,huck2019cross}, and question answering~\citep{liu2019xqa,lewis2019mlqa}. Recently, several methods have been proposed for continual learning~\citep{rusu2016progressive,kirkpatrick2017overcoming,lopez2017gradient,fernando2017pathnet,lee2017overcoming}, and these were applied to some NLP tasks, such as opinion mining~\citep{shu2016lifelong}, document classification~\citep{shu2017doc}, and dialogue state tracking~\citep{wu2019transferable}.

\section{Methodology}
\label{sec:methodology}
As shown in Figure \ref{fig:model}, our model consists of three parts: intra-modal encoder networks, emotion embedding mapping modules, and an inter-modal fusion module. In this section, we first define the problem, and then we introduce the details of our model. 

\subsection{Problem Definition}
\label{sec:problem_definition}
We define the input multi-modal data samples as $X=\{(t_i,a_i,v_i)\}_{i=1}^{I}$, in which $I$ denotes the total number of samples, and $t$, $a$, $v$ denote the \textit{textual}, \textit{acoustic}, and \textit{visual} modalities, respectively. For each modality, there is a set of emotion embeddings that represent the semantic meanings for the emotion categories to be recognized. In the textual modality, we have $E_t=\{e^t_{k}\}_{k=1}^{K}$, which is from the pre-trained GloVe embeddings. In acoustic and visual modalities, we have $E_a=\{e^a_{k}\}_{k=1}^{K}$ and $E_v=\{e^v_{k}\}_{k=1}^{K}$, which are mapped from $E_t$ by the mapping function \(f_{t \rightarrow v}\) and \(f_{t \rightarrow a}\). $K$ denotes the number of emotion categories, and it can be changed to fit different tasks and zero-shot learning. $Y=\{y_i\}_{i=1}^{I}$ denotes the annotations for  multi-label emotion recognition, where $y_i$ is a vector of length $K$ with binary values.

\subsection{Intra-modality Encoder Networks}
\label{sec:encoder}
As shown in Figure~\ref{fig:model}, for each data sample, there are three sequences of length $T$ from the three modalities. For each modality, we use a bi-directional Long-Short Term Memory (LSTM)~\citep{lstm} network as the encoder to process the sequence and get a vector representation. In other words, for the $i^{th}$ data sample, we will have three vectors, $r_t^{(i)} \in \mathbb{R}^{d_t}$, $r_a^{(i)} \in \mathbb{R}^{d_a}$, and $r_v^{(i)} \in \mathbb{R}^{d_v}$, that represent the \textit{textual}, \textit{acoustic}, and \textit{visual} modalities. Here, $d_t$, $d_a$, and $d_v$ are the dimensions of the emotion embeddings of the \textit{textual}, \textit{acoustic}, and \textit{visual} modalities, respectively. 

\subsection{Modality Mapping Module}
\label{sec:modality_mapping_module}
As mentioned in Section \ref{sec:introduction}, previous works do not consider the connections in different emotion categories, and the only information about emotions is in the annotations. In our model, we use emotion word embeddings to inject the semantic information of emotions into the model. Additionally, emotion embeddings also contain the relationships between emotion categories. For the textual modality, we use pre-trained GloVe~\citep{glove} embeddings of $K$ emotion words, denoted as $E_t \in \mathbb{R}^{K \times d_t}$. For the other two modalities, because there are no off-the-shelf pre-trained emotion embeddings, our model learns two mapping functions which project the vectors from the textual space into the acoustic and visual spaces:
\begin{align}
    E_a = f_{t \rightarrow a} (E_t) \in \mathbb{R}^{K \times d_a} \\
    E_v = f_{t \rightarrow v} (E_t) \in \mathbb{R}^{K \times d_v} \text{.}
\end{align}

\subsection{Modality Fusion and Prediction}
\label{sec:modality_fusion}
To predict the emotions for input sentences, we calculate the similarity scores between the sequence representation and the emotion embeddings for each modality. As shown in Eq.\ref{eq:distance}, for a data sample $i$, every modality will have a vector of similarity scores of length $K$ by dot product attention. We further add a modality fusion module to weighted sum all the vectors, in which the weights are also optimized end-to-end (Eq.\ref{eq:weighted_sum}). Finally, as the datasets are multi-labelled, the sigmoid activation function is applied to each score in the fused vector $s^{(i)}$, and a threshold is used to decide whether an emotion exists or not.
\begin{align}
    \label{eq:distance}
    &s_t^{(i)} = E_t r_t^{(i)}\text{, } s_a^{(i)} = E_a r_a^{(i)}\text{, } s_v^{(i)} = E_v r_v^{(i)}  \\
    \label{eq:weighted_sum}
    &s^{(i)} = \text{\textit{Sigmoid} } (w_t s_t^{(i)} + w_a s_a^{(i)} + w_v s_v^{(i)})
\end{align}

\section{Unseen Emotion Prediction}
\label{sec:unseen_emotion_prediction}
Collecting numerous training samples for a new emotion, especially for a low-resource emotion, is expensive and time-consuming. Therefore, in this section, we concentrate on the ability of our model to generalize to an unseen target emotion by considering the scenario where we have zero or only a few training samples in an unseen emotion. 

\subsection{Zero-Shot Emotion Prediction}
\label{sec:zero_shot_emotion_prediction}
Ideally, our model is able to directly adapt to a new emotion based on its embedding. Given a new text emotion embedding $e^t_{k+1}$, we can generate the visual and acoustic emotion embeddings $e^v_{k+1}$ and $e^a_{k+1}$, respectively, using the already learned mapping functions $f_{t \rightarrow v}$ and $f_{t \rightarrow a}$. After that, the similarity scores between the input sentence and the new emotion can be computed for each modality.

\subsection{Few-Shot Emotion Prediction}
\label{sec:few_shot_joint_training_and_continue_learning}
In this section, we assume 1\% of the positive training samples in a new emotion are available, and to balance the training samples, we take the same amount of negative training samples for the new emotion. However, the model could lose its ability to predict the original emotions when we simply fine-tune it on the training samples of a new emotion. To cope with this issue, we propose two fine-tuning settings. First, after we obtain the trained model in the source emotions, we jointly train it with the training samples of the source emotions and the new target emotion. Second, we utilize a continual learning method, gradient episodic memory (GEM)~\citep{lopez2017gradient}, to prevent the catastrophic forgetting of previously learned knowledge. The purpose of using continual learning is that we do not need to retrain with all the data from previously learned emotions since the data might not be available. We describe the training process of GEM as follows: 

We define $\Theta_S$ as the model's parameters trained in the source emotions, and $\Theta$ denotes the current optimized parameters based on the target emotion data. GEM keeps a small number of samples N from the source emotions, and a constraint is applied on the gradient to prevent the loss on the stored samples from increasing when the model learns the new target emotion.
The training process can be formulated as
\begin{align*}
    &\text{Minimize}_{\Theta}~~ L(\Theta) \\
    &\text{Subject to}~~ L(\Theta, N) \leq L(\Theta_S, N),
\end{align*}
where $L(\Theta, N)$ is the loss value of the N stored samples.

\section{Experiments}
\label{sec:experiments}

In this section, we first introduce the two public datasets we use and data feature extraction. Then, we discuss our evaluation metrics, including their advantages and defects. Finally, we introduce the baselines and our experimental settings. 

\subsection{Datasets}
\label{sec:datasets}

\paragraph{CMU-MOSEI} CMU Multimodal Opinion Sentiment and Emotion Intensity (CMU-MOSEI)~\citep{cmu-mosei} is currently the largest public dataset for multi-modal sentiment analysis and emotion recognition. It comprises 23,453 annotated data samples extracted from 3228 videos. For emotion recognition, it consists of six basic categories: \textit{anger}, \textit{disgust}, \textit{fear}, \textit{happy}, \textit{sad}, and \textit{surprise}. For zero-shot and few-shot learning evaluation, we use four relatively low-resource categories among them (\textit{anger}, \textit{disgust}, \textit{fear}, \textit{surprise}). The model is trained on the other five categories when evaluating one zero-shot category. A detailed statistical table about these categories is included in Appendix A.

\paragraph{IEMOCAP} The Interactive Emotional Dyadic Motion Capture (IEMOCAP)~\citep{iemocap} dataset was created for multi-modal human emotion analysis, and was collected from dialogues performed by ten actors. It is also a multi-labelled emotion recognition dataset which contains nine emotion categories. For comparison with prior works~\citep{RAVEN,rmfn, found-in-translation, MulT2019} where four (out of the nine) emotion categories are selected for training and evaluating the models, we also follow the same four categories, namely, \textit{happy}, \textit{sad}, \textit{angry}, and \textit{neutral}, to train our model. For zero-shot learning evaluation, we consider three low-resource categories from the remaining five, namely, \textit{excited}, \textit{surprised}, and \textit{frustrated}, as unseen emotions.

\subsection{Data Feature Extraction}
We use CMU-Multimodal SDK~\citep{cmu-sdk} for downloading and pre-processing the datasets. It helps to do data alignment and early-stage feature extraction for each modality. The textual data is tokenized in word level and represented using GloVe~\citep{glove} embeddings. Facial action units are extracted by the Facet~\citep{iMotions} to indicate muscle movements and expressions~\citep{Ekman1980FacialSO}. These are a commonly used type of feature for facial expression recognition~\citep{fan2020fau}. For acoustic data, COVAREP~\citep{Degottex2014COVAREPA} is used to extract fundamental features, such as mel-frequency cepstral coefficients (MFCCs), pitch tracking, glottal source parameters, etc.

\begin{table*}[t!]
\centering
\scalebox{0.73}{
\begin{tabular}{|l|cc|cc|cc|cc|cc|cc|cc|}
\hline
Emotion & \multicolumn{2}{c|}{Anger}    & \multicolumn{2}{c|}{Disgust}  & \multicolumn{2}{c|}{Fear}     & \multicolumn{2}{c|}{Happy}    & \multicolumn{2}{c|}{Sad}      & \multicolumn{2}{c|}{Surprise} & \multicolumn{2}{c|}{Average}  \\ \hline
Metrics & W-Acc          & AUC           & W-Acc          & AUC           & W-Acc          & AUC           & W-Acc          & AUC           & W-Acc          & AUC           & W-Acc          & AUC           & W-Acc          & AUC           \\ \hline
EF-LSTM & 58.5          & 62.2          & 59.9          & 63.9          & 50.1          & 69.8          & 65.1          & 68.9          & 55.1          & 58.6          & 50.6          & 54.3          & 56.6          & 63.0          \\
LF-LSTM & 57.7          & 66.5          & 61.0          & 71.9          & 50.7          & 61.1          & 63.9          & 68.6          & 54.3          & 59.6          & 51.4          & 61.5          & 56.5          & 64.9          \\
Graph-MFN     & 62.6          & -             & 69.1          & -             & 62.0          & -             & 66.3          & -             & 60.4          & -             & 53.7          & -             & 62.3          & -             \\
MTL     & 66.8          & 68.0$^\dagger$          & \textbf{72.7} & 76.7$^\dagger$          & 62.2          & 42.9$^\dagger$          & 53.6          & 71.4$^\dagger$          & 61.4          & 57.6$^\dagger$          & 60.6          & 65.1$^\dagger$          & 62.8          & 63.6$^\dagger$          \\ \hline
Ours    & \textbf{67.0} & \textbf{71.7} & 72.5          & \textbf{78.3} & \textbf{65.4} & \textbf{71.6} & \textbf{67.9} & \textbf{73.9} & \textbf{62.6} & \textbf{66.7} & \textbf{62.1} & \textbf{66.4} & \textbf{66.2} & \textbf{71.4} \\ \hline
\end{tabular}
}
\caption{Results of multi-modal emotion recognition on the CMU-MOSEI dataset. Baselines (EF-LSTM, LF-LSTM) and previous state-of-the-art models (Graph-MFN~\citep{cmu-mosei}, MTL~\citep{multi-task-senti-emo}) are compared. Results marked by $^\dagger$ are re-run and fine-tuned by us as they are not reported in the original paper.}
\label{tab:mosei_results}
\end{table*}

\begin{table*}[ht]
\centering
\scalebox{0.8}{
\begin{tabular}{|l|cc|cc|cc|cc|}
\hline
Emotion & \multicolumn{2}{c|}{Happy}    & \multicolumn{2}{c|}{Sad}      & \multicolumn{2}{c|}{Angry}    & \multicolumn{2}{c|}{Neutral}  \\ \hline
Metrics & Acc           & AUC           & Acc           & AUC           & Acc           & AUC           & Acc           & AUC           \\ \hline
EF-LSTM & 85.8          & 70.7          & 83.7          & 85.8          & 75.8          & 90.3          & 67.1          & 74.1          \\
LF-LSTM & 85.2          & 71.7          & 83.4          & 84.4          & 79.5          & 86.8          & 66.5          & 72.2          \\
RMFN~\citep{rmfn}    & \textbf{87.5} & -             & 83.8          & -             & 85.1          & -             & 69.5          & -             \\
RAVEN~\citep{RAVEN}   & 87.3          & -             & 83.4          & -             & 87.3          & -             & 69.7          & -             \\
MCTN~\citep{found-in-translation}    & 84.9          & -             & 80.5          & -             & 79.7          & -             & 62.3          & -             \\
MulT~\citep{MulT2019}    & 83.5$^\dagger$          & 71.2$^\dagger$          & 85.0$^\dagger$          & \textbf{89.3}$^\dagger$          & 85.5$^\dagger$          & 92.4$^\dagger$          & 71.0$^\dagger$          & \textbf{77.2}$^\dagger$          \\ \hline
Ours    & 85.0         & \textbf{74.2} & \textbf{86.6} & 88.4 & \textbf{88.1} & \textbf{93.2} & \textbf{71.1} & 76.7 \\ \hline
\end{tabular}
}
\caption{Multi-modal emotion recognition results on IEMOCAP. We re-run MulT  (marked by $^\dagger$) with its reported best hyper-parameters to get the AUC scores.}
\label{tab:iemocap_results}
\end{table*}

\subsection{Evaluation Metrics}
\label{sec:evaluation_metrics}

\paragraph{Weighted Accuracy} Due to the imbalanced nature of the emotion recognition dataset (for each emotion category, there are many more negative samples than positive samples), we use binary weighted accuracy~\citep{tong-2017-wacc} on each category to better measure the model's performance. The formula is 
\begin{align*}
    \text{Weighted Acc.} = \frac{TP \times N/P + TN}{2N}
\end{align*}

\noindent where P means total positive, TP true positive, N total
negative, and TN true negative.

\paragraph{Weighted F1}
In prior works~\citep{cmu-mosei,multi-task-senti-emo,MulT2019}, the binary weighted F1 score metric is used on the CMU-MOSEI dataset, and its formula is shown in Eq.\ref{eq:wf1}.
\begin{align}
    \text{Weighted F1} = \frac{P}{I} \times \text{F1}_p + \frac{N}{I} \times \text{F1}_n \label{eq:wf1}
\end{align}

\noindent Here, $\text{F1}_p$ is the F1 score that treats positive samples as \textit{positive}, while $\text{F1}_n$ treats negative samples as \textit{positive}, and they are weighted by their portion of the data. However, there is one defect of using binary weighted F1 in this task. As there are many more negative samples than positive ones, we find that with the increase of the threshold, the weighted F1 score will also increase because the \textit{true negative} increases. Therefore, in this paper, we do not report this metric. A detailed analysis of this is given in Appendix B. 

\paragraph{AUC Score} To eliminate the effect of threshold and mitigate the defect of the weighted F1 score, we also report Area under the ROC Curve (AUC) scores. The AUC score considers classification performance on both positive and negative samples, and it is scale- and threshold-invariant.

\subsection{Baselines}
\label{sec:baselines}
For both the CMU-MOSEI and IEMOCAP datasets, we use Early Fusion LSTM (EF-LSTM) and Late Fusion LSTM (LF-LSTM) as two baseline models. Additionally, for CMU-MOSEI, the Graph Memory Fusion Network (Graph-MFN)~\citep{cmu-mosei} and a multi-task learning (MTL) model~\citep{multi-task-senti-emo} are included for comparison with previous state-of-the-art models. For IEMOCAP, the Recurrent Multistage Fusion Network (RMFN)~\citep{rmfn}, Recurrent Attended Variation Embedding Network (RAVEN)~\citep{RAVEN}, and the Multimodal Transformer (MulT)~\citep{MulT2019} are included. 
To compare the AUC scores and zero-shot performance with baselines, we re-run the MTL and MulT models based on their reported best hyper-parameters, and we also carry out hyper-parameter search for a fair comparison.


\begin{table}[ht]
\centering
\scalebox{0.8}{
\begin{tabular}{|l|cc|}
\hline
 & CMU-MOSEI & IEMOCAP \\ \hline
Best Epoch & 15 & 16 \\
Batch size & 512 & 32 \\
Learning rate & 1e-4 & 1e-3 \\
\# LSTM layers & 2 & 2 \\
Hidden Size & 300/200/100 & 300/200/100 \\
Dropout & 0.15 & 0.15 \\
Gradient Clip & 10.0 & 1.0 \\
Random Seed & 0 & 0 \\ \hline
\end{tabular}
}
\caption{The hyper-parameters of our best models. The hidden size means the size of the LSTM hidden state of the textual/acoustic/visual modality, respectively. 
}
\label{tab:hyperparam}
\end{table}

\subsection{Training Details}
\label{sec:training_details}
The model is trained end-to-end with the Adam optimizer~\citep{adam} and a scheduler that will reduce the learning rate by a factor of 0.1 when the optimization stays on a plateau for more than 5 epochs.
The best hyper-parameters in our training for both datasets are shown in Table \ref{tab:hyperparam}. Also, we use the largest GloVe word embeddings (glove.840B.300d \footnote{\url{https://nlp.stanford.edu/projects/glove/}}) for both the input text data and the emotion embeddings in the textual modality. The weights of the textual embeddings are frozen during training to keep the pre-trained relations, which is also essential for doing zero-shot learning.

\section{Analysis}
\label{sec:analysis}

\begin{table*}[t!]
\centering
\scalebox{0.85}{
\begin{tabular}{|l|l|cc|cc|cc|cc|}
\hline
\multicolumn{2}{|l|}{Metrics} & W-Acc & AUC & W-Acc & AUC & W-Acc & AUC & W-Acc & AUC \\ \hline \hline
\multicolumn{2}{|l|}{Unseen emotion} & \multicolumn{2}{c|}{Anger (unseen)} & \multicolumn{2}{c|}{Disgust (unseen)} & \multicolumn{2}{c|}{Fear (unseen)} & \multicolumn{2}{c|}{Surprise (unseen)} \\ \hline
\multirow{3}{*}{Zero-Shot} & EF-LSTM & 50.6 & 50.9 & 50.3 & 48.2 & 45.8 & 42.3 & 50.2 & 46.9 \\
 & LF-LSTM & 48.4 & 49.2 & 49.7 & 44.2 & \textbf{47.4} & \textbf{47.3} & 48.6 & 48.3 \\
 & Ours & 55.9 & 61.6 & 67.5 & 72.7 & 41.8 & 40.6 & 53.4 & 55.5 \\ \hline
\multirow{3}{*}{1\% Few-Shot} & FT (Ours) & 58.9 & \textbf{61.9} & 67.9 & 71.5 & 43.1 & 43.1 & 51.8 & 53.9 \\
 & CL (Ours) & 58.9 & 61.5 & 68.7 & 72.8 & 42.6 & 42.7 & 50.6 & 52.5 \\
 & JT (Ours) & \textbf{59.0} & 61.1 & \textbf{69.2} & \textbf{74.2} & 41.9 & 41.7 & \textbf{55.2} & \textbf{58.1} \\ \hline \hline
 
\multicolumn{2}{|l|}{\begin{tabular}[c]{@{}l@{}}Average on all categories\end{tabular}} & \multicolumn{2}{c|}{Except Anger} & \multicolumn{2}{c|}{Except Disgust} & \multicolumn{2}{c|}{Except Fear} & \multicolumn{2}{c|}{Except Surprise} \\ \hline
Zero-shot & Ours & 65.6 & 70.6 & 64.4 & 69.3 & 65.9 & 70.9 & 67.2 & 71.4 \\ \hline
\multirow{3}{*}{1\% Few-Shot} & FT (Ours) & 64.4 & \textbf{69.8} & 63.7 & 68.5 & 65.4 & 70.7 & 65.1 & 71.4 \\
 & CL (Ours) & \textbf{64.6} & \textbf{69.8} & \textbf{63.8} & \textbf{68.9} & 65.6 & \textbf{70.9} & 65.5 & \textbf{71.5} \\
 & JT (Ours) & 64.3 & 69.3 & 63.5 & 68.8 & \textbf{65.9} & 70.8 & \textbf{66.1} & \textbf{71.5} \\ \hline
\end{tabular}
}
\caption{Zero/few-shot results on low-resource emotion categories in CMU-MOSEI dataset. Here, FT, CL, and JT stand for \textit{Fine-Tuning}, \textit{Continual Learning}, and \textit{Joint Training} respectively. FT directly fine-tunes the trained model on the unseen emotions, and CL and JT are two different settings introduced in Section \ref{sec:few_shot_joint_training_and_continue_learning}. Note that in the few-shot settings, we select the model based on the average performance of all emotions (including the unseen emotion) to ensure good overall performance of our model.}
\label{tab:mosei_zsl_fsl}
\end{table*}

\begin{figure*}[t!]
    \centering
    \includegraphics[width=\textwidth]{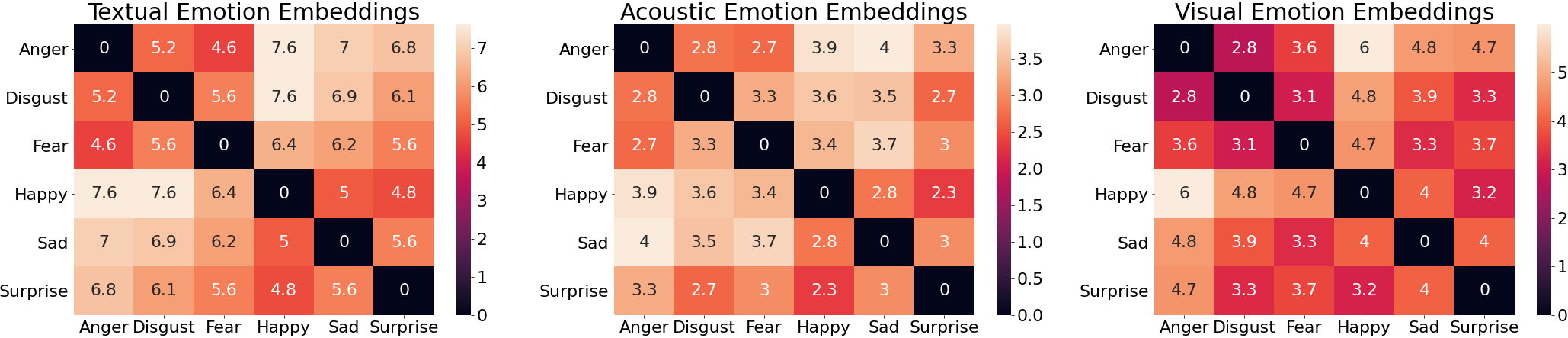}
    \caption{Euclidean distances between different emotion embeddings in the textual, acoustic, and visual spaces. Although the absolute values are different, the relative distances between emotion categories are well reserved. This indicates that the two mapping functions \(f_{t \rightarrow v}\) and \(f_{t \rightarrow a}\) transfer the relationships of emotion categories well.}
    \label{fig:heatmap}
\end{figure*}

\subsection{Results}
Table \ref{tab:mosei_results} shows our model's performance on the CMU-MOSEI dataset. Compared to existing baselines, our model surpasses them by a large margin. The weighted accuracy (W-Acc) and AUC score are used for evaluation, with a threshold set to 0.5 to calculate the W-Acc. As discussed in Section \ref{sec:evaluation_metrics}, we do not follow the previous papers in using the weighted F1-score (W-F1) because it does not provide an effective evaluation when the dataset is very imbalanced. For example, the weakest baseline, EF-LSTM, can even achieve ~90\% W-F1 by predicting almost all samples as negative. More plots and analysis of this defect of W-F1 are included in Appendix B. 

We further test our model on a second dataset called IEMOCAP, and the results are shown in Table \ref{tab:iemocap_results}. Similarly, our model achieves better results on most emotion categories, except \textit{happy}. For a fair comparison on IEMOCAP, we use accuracy instead of W-Acc, following the previous works compared in the table.

\subsection{Effects of Emotion Embeddings}
Quantitatively, our model makes a large improvement in the multi-modal emotion recognition task. We think it benefits greatly from the emotion embeddings, which can model the relationships (or distances) between emotion categories. This is especially important for emotion recognition, which is a multi-label task by nature, as people can have multiple emotions at the same time. For example, if a person is \textit{surprised}, it is more likely that this person is also \textit{happy} and \textit{excited} and is less likely to be \textit{disgusted} or \textit{sad}. This kind of information is expected to be modelled and captured by emotion embeddings. Intuitively, in the textual space, related emotions (e.g., \textit{angry} and \textit{disgusted}) tend to have closer word vectors than unrelated emotions (\textit{angry} and \textit{happy}). To ensure the effectiveness of word embeddings, for each emotion word, we investigated multiple forms of it. For example, for \textit{surprised}, we also tried with \textit{Surprised}, \textit{(S/s)urprising}, \textit{(S/s)urprise}. Generally, they all show a similar trend, and in most cases, the word form that is used to describe human shows the best results. In our final setting, we iterate and pick the best performing form for each emotion category.

Moreover, our model can transfer the relationship of emotion categories from the textual space to the acoustic and visual spaces using end-to-end optimized mapping functions. In Figure \ref{fig:heatmap}, we show the Euclidean distances of emotion embeddings between categories. The relative positions are preserved very well after being transferred from the textual space to the visual and acoustic spaces. This indicates that the learned mapping functions (\(f_{t \rightarrow v}\) and \(f_{t \rightarrow a}\)) are effective. Although it is not the main focus of this paper, we think improving the pre-trained textual emotion embeddings is an essential direction for future work. It can benefit all modalities and further enhance the overall performance. For example, incorporate semantic emotion information~\citep{xu2018emo2vec} to the original word embeddings.

\subsection{Zero/Few-Shot Results}
Benefiting from the pre-trained textual emotion embeddings and learned mapping functions, our model can recognize unseen emotion categories to a certain extent. We evaluate our model's zero-shot learning ability on the low-resource categories in CMU-MOSEI (shown in Table \ref{tab:mosei_zsl_fsl}) and IEMOCAP (shown in Table \ref{tab:iemocap_zsl_results}). For a fair comparison, we use the same training setting that is used in Table \ref{tab:mosei_results}. This can ensure that no downgrade happens on the seen emotions, and the model is not selected to overfit a single unseen category. As we can see, the zero-shot results of the baselines are similar to random guesses, because the weights related to that unseen emotion in the model are randomly initialized and have never been optimized. For our model, the zero-shot performance is much better than that of the baselines in almost all emotions. This is because our model learns to classify emotion categories based on the similarity between the sentence representation and emotion embeddings, which enables our model better generalization ability to other unseen emotions since emotion embeddings contain semantic information in the vector space.

Furthermore, we perform few-shot learning using only 1\% of data of these low-resource categories. As we can see from Table~\ref{tab:mosei_zsl_fsl}, using very few training samples, our model can adapt to unseen emotions without losing the performance in the source emotions. In addition, we observe that simply fine-tuning (FT) our trained model sometimes obtains inferior performance. This is because our model will gradually lose the ability to classify the source emotions, and we have to early stop the fine-tuning process, which leads to inferior performance. We can see that CL and JT prevent our model from catastrophic forgetting and improve the few-shot performance in the unseen emotion. Moreover, JT achieves slightly better performance than CL. This can be attributed to the fact that CL might still result in performance drops in source emotions since our model only observes partial samples from them. At the same time, JT directly optimizes the model on the data samples of such emotions.

\subsection{Ablation Study}
\label{sec:ablation_study}

\begin{table}[t]
\centering
\scalebox{0.79}{
\begin{tabular}{|l|cl|cl|cl|}
\hline
Unseen emotion & \multicolumn{2}{c|}{\begin{tabular}[c]{@{}c@{}}Excited\\ (unseen)\end{tabular}} & \multicolumn{2}{c|}{\begin{tabular}[c]{@{}c@{}}Surprised\\ (unseen)\end{tabular}} & \multicolumn{2}{c|}{\begin{tabular}[c]{@{}c@{}}Frustrated\\ (unseen)\end{tabular}} \\ \hline
Metrics & Acc & \multicolumn{1}{c|}{F1} & Acc & \multicolumn{1}{c|}{F1} & Acc & \multicolumn{1}{c|}{F1} \\ \hline
EF-LSTM & \multicolumn{1}{l}{13.1} & 23.1 & \multicolumn{1}{l}{11.3} & 5.1 & \multicolumn{1}{l}{22.9} & 37.2 \\
LF-LSTM & \multicolumn{1}{l}{14.0} & 23.3 & \multicolumn{1}{l}{2.6} & 5.1 & \multicolumn{1}{l}{23.7} & 37.4 \\
MulT & \multicolumn{1}{l}{45.1} & 27.3 & \multicolumn{1}{l}{41.4} & 7.5 & \multicolumn{1}{l}{48.7} & 40.9 \\
\hline
Ours (TAV) & 82.0 & \multicolumn{1}{c|}{56.1} & 78.8 & \multicolumn{1}{c|}{13.1} & 73.6 & \multicolumn{1}{c|}{57.9} \\
Ours (TA) & \multicolumn{1}{l}{79.9} & 52.7 & \multicolumn{1}{l}{79.3} & 14.4 & \multicolumn{1}{l}{75.1} & 60.1 \\
Ours (TV) & \multicolumn{1}{l}{75.9} & 42.7 & \multicolumn{1}{l}{58.6} & 9.1 & \multicolumn{1}{l}{54.1} & 13.6 \\
Ours (AV) & \multicolumn{1}{l}{89.1} & 69.9 & \multicolumn{1}{l}{65.7} & 13.2 & \multicolumn{1}{l}{83.9} & 73.6 \\
Ours (T) & 72.9 & \multicolumn{1}{c|}{37.1} & 67.7 & \multicolumn{1}{c|}{3.1} & 55.3 & \multicolumn{1}{c|}{9.0} \\
Ours (A) & 76.9 & \multicolumn{1}{c|}{52.8} & 82.1 & \multicolumn{1}{c|}{16.8} & 86.1 & \multicolumn{1}{c|}{74.8} \\
Ours (V) & 82.1 & \multicolumn{1}{c|}{35.0} & 81.1 & \multicolumn{1}{c|}{6.2} & 68.6 & \multicolumn{1}{c|}{44.6} \\ \hline
\end{tabular}
}
\caption{Zero-shot results on the IEMOCAP dataset. T (textual), A (acoustic), and V (visual) indicate the existence of that modality during inference time.}
\label{tab:iemocap_zsl_results}
\end{table}

\begin{table}[t]
\centering
\scalebox{0.75}{
\begin{tabular}{|l|cccccc|}
\hline
Metric & \multicolumn{6}{c|}{W-Acc} \\ \hline
Emotion & \multicolumn{1}{c}{Anger} & \multicolumn{1}{c}{Disgust} & \multicolumn{1}{c}{Fear} & \multicolumn{1}{l}{Happy} & \multicolumn{1}{l}{Sad} & Surprise \\ \hline
T+A+V & \textbf{67.0} & \textbf{72.5} & 65.4 & \textbf{67.9} & 62.6 & \textbf{62.1} \\
T+A & 65.0 & 71.9 & 64.8 & 66.0 & \textbf{63.0} & 59.9 \\
T+V & 64.9 & 71.2 & \textbf{66.7} & 67.6 & 61.0 & 60.4 \\
A+V & 63.8 & 71.1 & 65.5 & 64.5 & 61.3 & 55.2 \\
Only T & 61.5 & 69.0 & 64.3 & 64.2 & 59.7 & 61.2 \\
Only A & 61.9 & 71.5 & 66.9 & 62.7 & 61.0 & 54.8 \\
Only V & 63.4 & 69.7 & 63.2 & 63.2 & 58.5 & 53.3 \\ \hline
\end{tabular}
}
\caption{Ablation study on CMU-MOSEI dataset. Different combinations of subsets of modalities are used.}
\label{tab:mosei_ablation}
\end{table}

To further investigate how each individual modality influences the model, we perform comprehensive ablation studies on supervised multi-modal emotion recognition and also zero-shot prediction.

In Table \ref{tab:mosei_ablation}, we enumerate different subsets of the (\textit{textual}, \textit{acoustic}, \textit{visual}) modalities to evaluate the effect of each single modality. Generally, the performance will increase if more modalities are available. Compared to single-modal data, multi-modal data can provide supplementary information, which leads to more accurate emotion recognition. In terms of a single modality, we find that \textit{textual} and \textit{acoustic} are more effective than \textit{visual}. 
 
Similarly, in Table \ref{tab:iemocap_zsl_results}, we show the zero-shot performance with different combinations of modalities during the inference time (all modalities exist in the training phase). As there are many more negative samples than positive ones in the ZSL setting, we also evaluate the models with the unweighted F1 score. Because if a model has high accuracy but a low F1, it is heavily biased to the negative samples so it cannot do classification effectively. Empirical results indicate that zero-shot on only one modality is possible. Moreover, if the data of an emotion category has strong characteristics in one modality and is ambiguous in other modalities, single-modality can even surpass multi-modality on zero-shot prediction. For example, the performance of single-modality zero-shot prediction using the \textit{acoustic} modality on the \textit{surprised} category is better than using all modalities. 

\section{Conclusion}
\label{sec:conclusion}
In this paper, we introduce a modality-transferable model that leverages cross-modality emotion embeddings for multi-modal emotion recognition. It makes predictions by measuring the distances between input data and target emotion categories, which is especially effective for a multi-label problem. The model also learns two mapping functions to transfer pre-trained textual emotion embeddings to acoustic and visual spaces. The empirical results demonstrate that it exhibits state-of-the-art performance on most of the categories. Enabled by the utilization of emotion embeddings, our model can carry out zero-shot learning for unseen emotion categories and can quickly adapt few-shot learning without downgrading trained categories. 

\section*{Acknowledgement}
This work is funded by MRP/055/18 of the Innovation Technology Commission, the Hong Kong SAR Government.

\bibliography{aacl-ijcnlp2020}
\bibliographystyle{acl_natbib}

\appendix

\section{Statistics of Datasets}
\label{appendix:A}

\begin{table}[h!]
\centering
\scalebox{0.9}{
\begin{tabular}{|l|lll|}
\hline
 & Train & Valid & Test \\ \hline \hline
Anger & 3443 & 427 & 971 \\
Disgust & 2720 & 352 & 922 \\
Fear & 1319 & 186 & 332 \\
Happy & 8147 & 1313 & 2522 \\
Sad & 3906 & 576 & 1334 \\
Surprise & 1562 & 201 & 479 \\ \hline
\end{tabular}
}
\caption{Statistics of the CMU-MOSEI dataset. Some emotion categories are very low-resource.}

\label{tab:mosei_stats}
\end{table}

\begin{table}[h!]
\centering
\scalebox{0.9}{
\begin{tabular}{|l|ccc|}
\hline
 & \multicolumn{1}{l}{Train} & \multicolumn{1}{l}{Valid} & \multicolumn{1}{l|}{Test} \\ \hline \hline
Happy & 338 & 116 & 135 \\
Sad & 690 & 188 & 193 \\
Angry & 735 & 136 & 227 \\
Neutral & 954 & 358 & 383 \\ \hline \hline
Excited & - & - & 141 \\
Surprised & - & - & 25 \\
Frustrated & - & - & 278 \\ \hline
\end{tabular}
}
\caption{Statistics of emotion categories in the IEMOCAP dataset. The three at the bottom are unseen emotions used for the evaluation of zero-shot learning.}
\label{tab:iemocap_stats}
\end{table}

\section{Weighted F1 Analysis}
\label{appendix:B}

\begin{figure}[!ht]
    \centering
    \includegraphics[width=\linewidth]{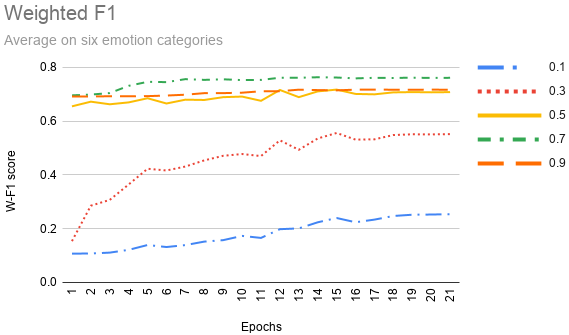}
    \caption{Trend lines of the weighted f1 (W-F1) score during training on the validation set of CMU-MOSEI.}
    \label{fig:app-B-1}
\end{figure}

\begin{figure}[!ht]
    \centering
    \includegraphics[width=\linewidth]{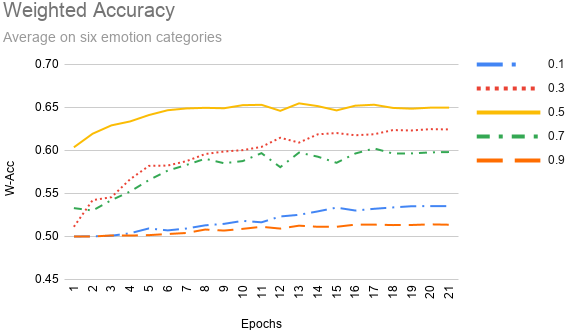}
    \caption{Trend lines of the weighted accuracy (W-Acc) score during training on the validation set of CMU-MOSEI.}
    \label{fig:app-B-2}
\end{figure}

In Figure \ref{fig:app-B-1} and \ref{fig:app-B-2}, we show the trends of the weighted F1 (W-F1) and weighted accuracy (W-Acc) on the validation set of CMU-MOSEI during the training phase. The lines represent different threshold values as shown in the legend of each figure. The W-F1 is almost proportional to the threshold values and it is still very high when the threshold is 0.9 (i.e. most data samples are predicted to be negative). Moreover, when the threshold is large, the W-F1 keeps a high value starting from epoch 1. By contrast, the W-Acc score is more reliable. It ensures the model can also retrieve positive samples. We observe a similar phenomenon on all models. As a result, we think W-F1 is unsuitable as an evaluation metric on this dataset.

\end{document}